\documentclass[10pt,twocolumn,letterpaper]{article}

\usepackage[pagenumbers]{iccv} 

\usepackage{afterpage}

\usepackage{pgfplots}
\pgfplotsset{compat=1.18}
\usepackage{pifont}
\usepackage{graphicx}
\usepackage{makecell}
\usepackage{amsmath}
\usepackage{amssymb}
\usepackage{booktabs}
\usepackage{multirow}
\usepackage{cuted}
\newcolumntype{a}{>{\columncolor{Gray}}c}
\definecolor{iccvblue}{rgb}{0.21,0.49,0.74}
\usepackage[pagebackref,breaklinks,colorlinks,allcolors=iccvblue]{hyperref}

\def\paperID{3947} 
\def\confName{ICCV}
\def\confYear{2025}

\title{GaussianFlowOcc: Sparse and Weakly Supervised Occupancy Estimation using Gaussian Splatting and Temporal Flow}

\author{Simon Boeder\\
Robert Bosch GmbH\\
{\tt\small simon.boeder@de.bosch.com}
\and
Fabian  Gigengack\\
Robert Bosch GmbH\\
{\tt\small fabian.gigengack@de.bosch.com}
\and
Benjamin Risse\\
University of M\"unster\\
{\tt\small b.risse@uni-muenster.de}
}

\begin{document}
\maketitle

\begin{strip}
    \vspace{-1cm}
    \centering
    \includegraphics[page=1, width=\textwidth]{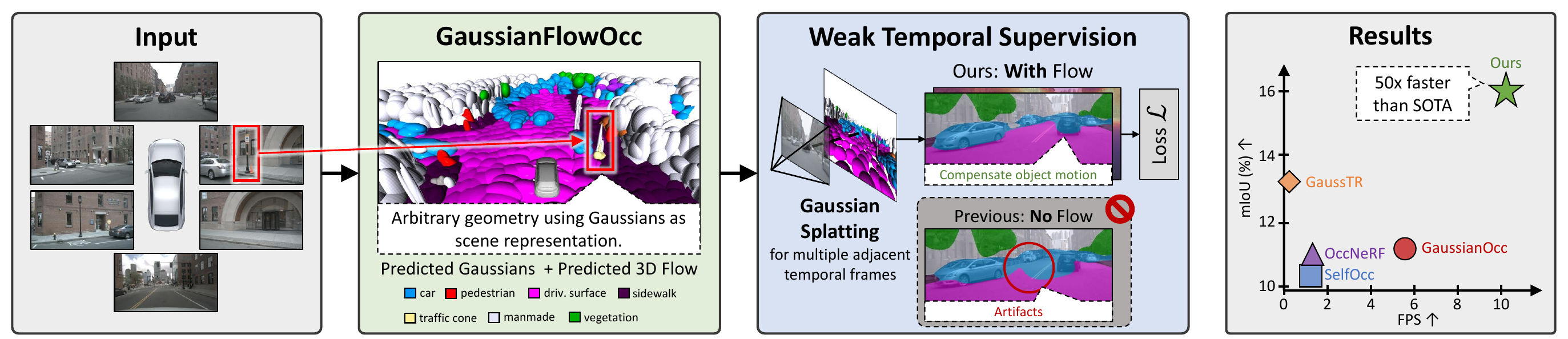}
    \captionof{figure}{
        \textbf{Contributions of GaussianFlowOcc.} 
        We propose an efficient and fast model for estimating dynamic scenes using 3D Gaussian distributions, trained with weak supervision.
        By employing a \emph{Temporal Module}, we efficiently model scene dynamics to account for object motion during training.
        Our approach significantly outperforms previous methods in terms of mIoU and inference speed.
    }
    \label{fig:teaser}
\end{strip}

\begin{abstract}
Occupancy estimation has become a prominent task in 3D computer vision, particularly within the autonomous driving community.
In this paper, we present a novel approach to occupancy estimation, termed \emph{GaussianFlowOcc}, which is inspired by Gaussian Splatting and replaces traditional dense voxel grids with a sparse 3D Gaussian representation.
Our efficient model architecture based on a \emph{Gaussian Transformer} significantly reduces computational and memory requirements by eliminating the need for expensive 3D convolutions used with inefficient voxel-based representations that predominantly represent empty 3D spaces.
\emph{GaussianFlowOcc} effectively captures scene dynamics by estimating temporal flow for each Gaussian during the overall network training process, offering a straightforward solution to a complex problem that is often neglected by existing methods.
Moreover, \emph{GaussianFlowOcc} is designed for scalability, as it employs weak supervision and does not require costly dense 3D voxel annotations based on additional data (e.g., LiDAR).
Through extensive experimentation, we demonstrate that \emph{GaussianFlowOcc} significantly outperforms all previous methods for weakly supervised occupancy estimation on the nuScenes dataset while featuring an inference speed that is 50 times faster than current SOTA.
\end{abstract}

\vspace{-5mm}
\section{Introduction}
Vision-based 3D object detection and occupancy estimation are fundamental challenges in the field of autonomous driving, as an accurate understanding of a vehicle's surroundings is essential for reliable planning and navigation~\cite{xu2025survey, shi2024grid, zhang2024vision}.
Despite recent advances, existing approaches face several limitations.
First, most recent occupancy estimation models rely on fully annotated 3D ground truth data for training, which is expensive and challenging to obtain at scale.
Second, these models depend on dense 3D voxel-based representations, leading to substantial computational overhead that is often unnecessary, given the natural sparsity of real-world 3D scenes.
Although self-supervised methods have been developed to reduce reliance on 3D labels by leveraging temporal rendering~\cite{huang2023selfocc, zhang2023occnerf}, these approaches fail to account for the inherent dynamics of driving scenarios. 
This introduces temporal inconsistencies during training, severely limiting their performance.

To address these limitations, we introduce \emph{GaussianFlowOcc}, a novel occupancy estimation method that combines efficiency and expressiveness through several contributions.
At the core of our approach is a sparse representation of the scene that utilizes 3D Gaussian distributions, replacing traditional dense voxel grids.
We propose a \emph{Gaussian Transformer} network, a novel deep learning model designed to directly transform image features into 3D Gaussians.
This shift not only reduces computational overhead but also provides a more flexible representation for modeling dynamic scenes.
Additionally, it allows us to leverage \emph{Gaussian Splatting}~\cite{kerbl20233d} for efficient training without relying on 3D voxel labels.
The estimated Gaussians are rendered back into the input camera space, where losses can be computed using 2D labels such as pseudo ground truth derived from pretrained models like Grounded-SAM~\cite{ren2024grounded} for semantics and Metric3D~\cite{yin2023metric3d} for depth.
Another key contribution is \emph{Temporal Gaussian Splatting}, where adjacent frames are rendered to enhance the supervisory signal, allowing the model to capture 3D geometry more effectively.
To handle temporal inconsistencies caused by dynamic objects, GaussianFlowOcc extends this framework with a \emph{Temporal Module} that estimates a 3D flow for each Gaussian to each rendered temporal frame.
This additional modeling of scene dynamics mitigates errors caused by object motion during training, strengthening the supervision signal and improving the overall performance.
By combining these techniques, GaussianFlowOcc achieves state-of-the-art performance while maintaining computational efficiency, rendering it highly effective for practical occupancy estimation applications.
In summary, our contributions are:
\begin{itemize}
    \item \textbf{Efficient Gaussian Transformer for sparse 3D Gaussian scene representation:} 
    We introduce a novel \emph{Gaussian Transformer} that leverages set-based attention mechanisms to efficiently estimate a sparse set of 3D Gaussian distributions representing the scene from multi-view images.
    It entirely eliminates voxelization and computationally intensive operations, leading to an inference speedup of $50\times$ compared to state-of-the-art.
    At the same time, by leveraging Gaussian Splatting, our method learns from 2D pseudo labels instead of relying on costly 3D annotations from other sensors like LiDAR.
    Our approach requires only images for training, making it more scalable than methods depending on 3D labeled datasets.
    \item \textbf{Modeling of scene dynamics via learned 3D flow:} GaussianFlowOcc integrates a \emph{Temporal Module} that additionally estimates 3D flow for each Gaussian, enabling robust handling of dynamic objects during inference and during \emph{Temporal Gaussian Splatting} at training time.
    GaussianFlowOcc is the first method to incorporate temporal supervision while explicitly modeling scene dynamics, without requiring ground truth flow annotations.
    \item \textbf{State-of-the-art performance:} On top of being highly efficient at inference, GaussianFlowOcc substantially outperforms previous weakly supervised methods.
\end{itemize}

\section{Related Work}

\subsection{Occupancy Estimation}
The 3D semantic occupancy estimation task has become an important area in the autonomous driving research community in recent years.
Numerous voxel-based occupancy benchmarks have been introduced for datasets such as SemanticKITTI~\cite{behley2019semantickitti} and nuScenes~\cite{tian2023occ3d, wang2023openoccupancy}.
Early works in 3D occupancy estimation utilize established techniques from Birds-Eye-View (BEV) perception and object detection \cite{huang2021bevdet,li2022bevformer} to lift multi-view camera images into a unified 3D voxel grid~\cite{huang2023tri, huang2022bevdet4d, tong2023scene, cao2022monoscene, li2023voxformer}.
Subsequent approaches have improved model efficiency \cite{yu2023flashocc, wang2024opus, lu2023octreeocc, liu2024fully, shi2025occupancy, tang2024sparseocc}, optimized the training procedure and label efficiency \cite{pan2023renderocc, boeder2024occflownet, gan2023simple, hayler2024s4c, sun2024gsrender}, and improved occupancy estimation performance through architectural innovations \cite{li2023fb, zhang2023occformer, jiang2023symphonize, tan2024geocc, Zhao_2024_CVPR, ma2024cotr, ma2024cam4docc}.
Specifically, GaussianFormer \cite{huang2024gaussianformer} and GaussianWorld \cite{zuo2025gaussianworld} employ a sparse and efficient Gaussian representation similar to our work.

Despite these advancements, the majority of existing 3D occupancy estimation methods require costly voxel-based 3D ground truth labels for training.
Therefore, a parallel line of research has emerged that focuses on self and weakly supervised learning for occupancy estimation models, leveraging only 2D labels.
SelfOcc~\cite{huang2023selfocc} and OccNeRF~\cite{zhang2023occnerf} employ volume rendering inspired by NeRF~\cite{mildenhall2021nerf} to render estimated 3D occupancy back to the 2D image space. 
Photometric consistency losses can be used to train the geometry estimation, while semantic information is incorporated using pretrained foundation models like OpenSeeD~\cite{zhang2023simple} or GroundedSAM~\cite{ren2024grounded}.
GaussianOcc~\cite{gan2024gaussianocc} replaces the volume rendering pipeline with 3D Gaussian Splatting to accelerate training, yet it continues to model the scene with dense occupancy grids, thus failing to exploit the benefits of a fully sparse Gaussian representation.
Also, while these methods eliminate the need for 3D labels, they fail to address scene dynamics, a critical aspect when relying on temporal consistency losses.

A growing body of research aims to align 3D occupancy with feature spaces of strong foundation models.
OccFeat~\cite{sirko2024occfeat} distills features of CLIP~\cite{radford2021learning} and DINO~\cite{caron2021emerging, oquab2023dinov2} into an occupancy representation for model pretraining.
POP-3D~\cite{vobecky2024pop}, LOcc~\cite{yu2024language} and OVO~\cite{tan2023ovo} perform open-vocabulary occupancy estimation by aligning voxel-based predictions with vision-language features extracted from pretrained encoders like MaskCLIP~\cite{zhou2022extract}.
VEON~\cite{zheng2025veon} and LangOcc~\cite{boeder2024langocc} follow up on the self-supervised methods and directly use volume rendering of vision-language features to train open-vocabulary models.

\subsection{Differentiable Rendering and 3D Gaussian Splatting}
Differentiable rendering has emerged as a powerful technique for learning 3D scene representations by projecting them into 2D views, followed by an optimization based on photometric or semantic consistency.
Neural Radiance Fields (NeRF)~\cite{mildenhall2021nerf} have been particularly influential, modeling scenes as volumetric representations that encode radiance and density, enabling novel view synthesis through differentiable volume rendering.
Recently, 3D Gaussian Splatting (GS)~\cite{kerbl20233d} has introduced a novel paradigm for 3D scene reconstruction by representing scenes as a collection of 3D Gaussians.
This approach significantly reduces computational overhead while preserving expressive scene modeling, making it highly efficient for dynamic or large-scale reconstructions.
NeRF and GS approaches were originally designed to reconstruct individual scenes for novel-view synthesis, with research focusing on improving efficiency~\cite{muller2022instant, chen2025mvsplat}, rendering quality~\cite{barron2021mip, barron2022mip} or feature enrichment~\cite{qin2024langsplat, kerr2023lerf, ye2023featurenerf, zhou2024feature}.
Several works have further explored the modeling of dynamic scenes for video reconstruction \cite{wu20244d, fridovich2023k, pumarola2021d}.
Another line of work incorporates different priors like depth, stereo-matching, or additional data like LiDAR, to train generalizable reconstruction models~\cite{xu2022point,chang2022rc,chen2021mvsnerf,yu2021pixelnerf, wimbauer2023behind, liu2025mvsgaussian, zheng2024gps}.
Several methods have also been developed specifically for reconstruction within autonomous driving scenarios~\cite{lu2024drivingrecon,zhou2024drivinggaussian,yan2024street}, while GaussianAD~\cite{zheng2024gaussianad} employs a 3D Gaussian representation for end-to-end driving.
As mentioned above, similar ideas have been adapted to train occupancy estimation models.
Methods like OccNeRF~\cite{zhang2023occnerf}, SelfOcc~\cite{huang2023selfocc} and GaussianOcc~\cite{gan2024gaussianocc} employ volume rendering or GS for weakly supervised training.
However, they still rely on voxel grids for scene representation, which limits efficiency and scalability.
GSRender~\cite{sun2024gsrender} additionally introduces a ray compensation method to mitigate duplicate predictions along camera rays.
Lastly, the recent approach GaussTR~\cite{jiang2024gausstr} shares similarities with our method by adopting 3D Gaussians as the scene representation.
However, GaussTR uses multiple pretrained feature encoders (e.g., CLIP, Metric3D, SAM) during inference, making the pipeline computationally expensive.
Furthermore, GaussTR employs standard attention layers, which significantly limits the number of Gaussians it can handle. 

\section{Methodology}\label{sec:methodology}
\subsection{Problem Definition and Scene Representation}\label{sec:problem_definition}
The objective of occupancy prediction is to accurately estimate the 3D geometry and semantics surrounding a vehicle based on a set of $L$ multi-view images $I = \{I^1, I^2, ..., I^L\}$ at the current time step.
Previous methods have typically utilized a semantic voxel volume $V=\{c_1, c_2, ... c_{C}\}^{X \times Y \times Z}$ on a predefined grid, with $C$ representing the number of semantic classes, as a structured representation of the scene.
In contrast, our approach involves estimating the 3D scene using 3D Gaussian distributions, drawing inspiration from 3D Gaussian Splatting~\cite{kerbl20233d}.
The scene is defined as a collection of $N$ Gaussians $\mathcal{G} = \{G_1, G_2, ..., G_N\}$, each characterized by a set of properties: mean $\mu \in \mathbb{R}^3$, opacity $\sigma \in [0,1]$, scale $s \in \mathbb{R}^{3}$, rotation quaternions $r \in \mathbb{R}^4$, and semantic logits $c \in \mathbb{R}^{C}$:
\begin{align}
G_i = (\mu, \sigma, s, r, c).
\label{eq:scene_definition}
\end{align}
We train a model $\mathbb{M}(I)$ to estimate the 3D Gaussians $\mathcal{G}$.
If desired, the resulting scene representation can be easily converted to a voxel volume as a post-processing step to facilitate comparison with previous methods (see \cref{sec:voxelize}).

\subsection{Model Architecture}\label{sec:model_arch}
\begin{figure*}
    \centering
    \includegraphics[page=1, trim=0cm 8.86cm 3.68cm 0cm, clip, width=\textwidth]{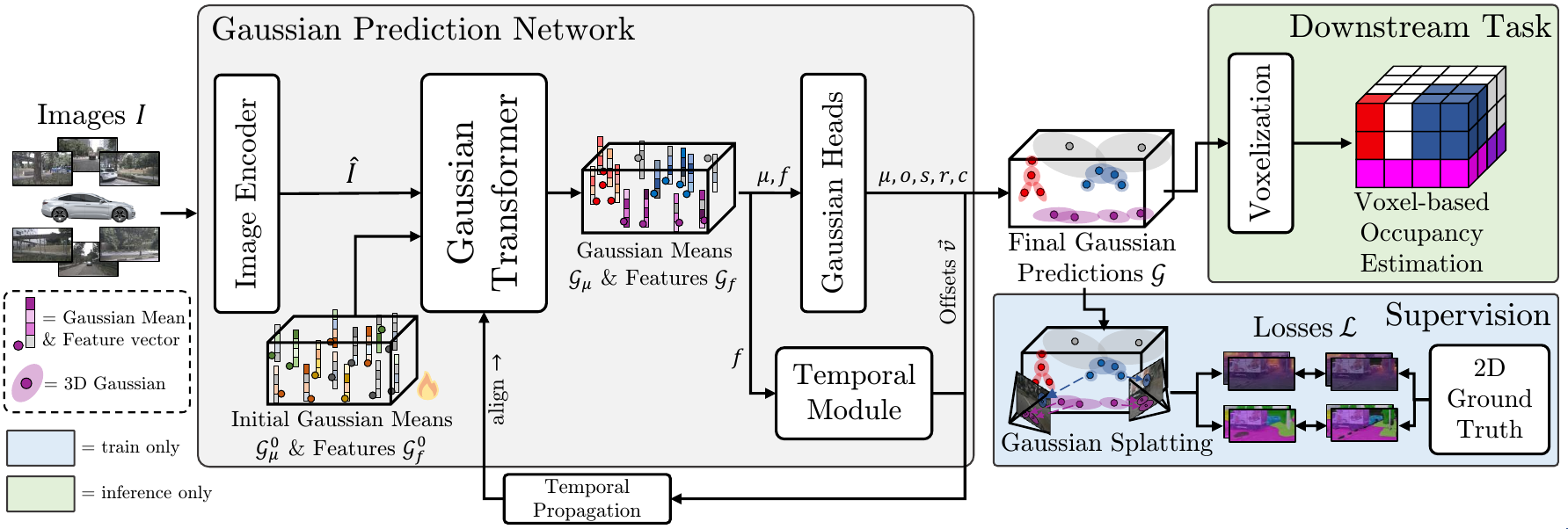}
    \caption{
        \textbf{Overview of GaussianFlowOcc.}
        After encoding a set of input images, the \emph{Gaussian Transformer} iteratively estimates the position and features of 3D Gaussian distributions.
        The \emph{Gaussian Heads} then predict opacity, scale, rotation and semantics of these Gaussians.
        The model is trained using Gaussian Splatting with 2D labels, generated from off-the-shelf models.
        The \emph{Temporal Module} simultaneously estimates 3D temporal offsets for each Gaussian to correct temporal inconsistencies when using \emph{Temporal Gaussian Splatting}.
    }
    \label{fig:architecture}
\end{figure*}

The architecture of the proposed model is illustrated in \cref{fig:architecture}.
Initially, image features $\hat{I}$ are extracted from the input images using an image encoder.
The image features, along with the initial positions and features of the Gaussians, are then fed into the proposed \emph{Gaussian Transformer} (\cref{sec:gaussian_transformer}).
The \emph{Gaussian Transformer} iteratively transforms the initial Gaussians to their final positions by employing cross-attention to the image features, self-attention between the Gaussians, and temporal attention to the Gaussians of the previous frame. 
Subsequently, the \emph{Gaussian Heads} (\cref{sec:gaussian_head}) estimate the remaining Gaussian properties and their semantics.
Concurrently, the \emph{Temporal Module} (\cref{sec:gaussian_flow}) computes a 3D flow for each Gaussian to its temporal neighbors.
The final Gaussians and their temporal offsets are then utilized in the Gaussian Splatting pipeline (\cref{sec:gaussian_splatting}) to render depth and semantic maps into the input cameras and a set of temporally adjacent frames.
The model is trained using losses between the rendered and ground truth depth and semantic segmentation maps.

\subsubsection{Initial Gaussians} \label{sec:initial}
We initialize with a set of $N$ Gaussian means $\mathcal{G}^0_\mu \in \mathbb{R}^{N \times 3}$ and latent features $\mathcal{G}^0_f \in \mathbb{R}^{N \times D}$, akin to the queries in previous works \cite{li2022bevformer, liu2022petr,wang2023exploring}.
The initial positions and features of the Gaussians are learnable, enabling the model to incorporate prior knowledge of driving scenes.
Notably, the other Gaussian properties $o, s, r, c$ are left uninitialized; instead, the model learns to infer them from the Gaussian features through the \emph{Gaussian Heads} in the output stage, a strategy that has proven to be more robust for training.

\subsubsection{Image Encoder} \label{sec:image encoder}
Image features $\hat{I}$ are extracted from the input images $I$ using a pretrained backbone architecture like \textit{ResNet-50}~\cite{he2016deep}.
These features and the initial Gaussians are then fed into the \emph{Gaussian Transformer}.

\subsubsection{Gaussian Transformer} \label{sec:gaussian_transformer}
\begin{figure}
    \centering
     \resizebox{\linewidth}{!}{\includegraphics[page=2, trim=0cm 4.08cm 6.9cm 0cm, clip]{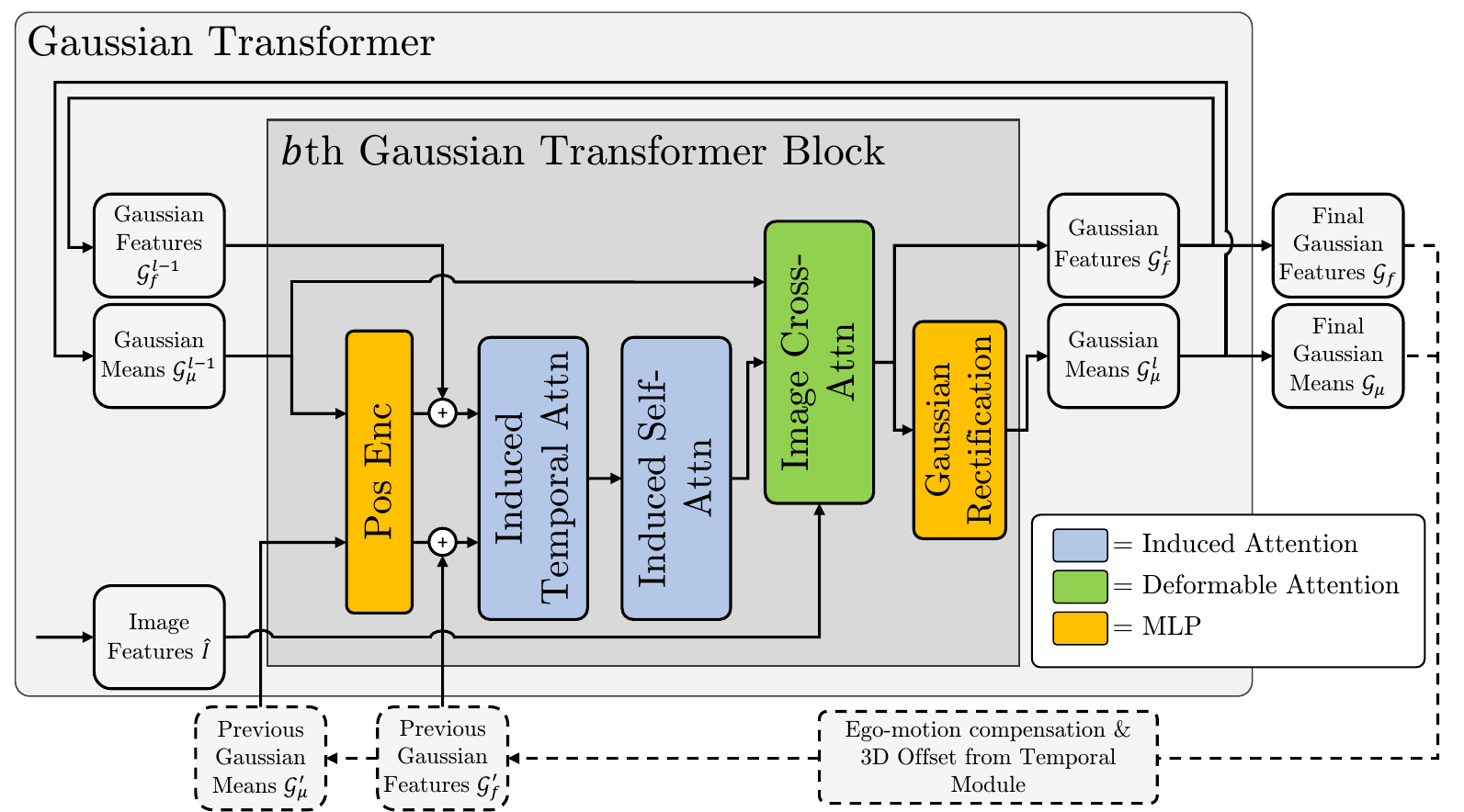}}
    \caption{
        \textbf{Architecture of the proposed \emph{Gaussian Transformer}.}
        In each block, the features and means of the Gaussians are refined using Pos Enc, ITA, ISA, GICA and Gaussian Rectification.
    }
    \label{fig:gauss_transformer}
    \vspace{-3mm}
\end{figure}

The \emph{Gaussian Transformer} iteratively updates the initial Gaussian means and features over $B$ blocks, with each block $b$ comprising five successive modules, elaborated in the subsequent sections.
An overview of a \emph{Gaussian Transformer} block is depicted in \cref{fig:gauss_transformer}. 

\paragraph{Positional Encoding}
The Gaussian means of the previous block $\mathcal{G}_\mu^{b-1}$ are encoded into the latent dimension $D$ using a Multi-Layer Perceptron (MLP) and are added onto the previous features $\mathcal{G}_f^{b-1}$.

\paragraph{Gaussian-Image Cross-Attention}
To enable Gaussians to acquire meaningful scene information, we introduce the \emph{Gaussian-Image Cross-Attention} (GICA), which facilitates interactions between Gaussians $\mathcal{G}$ and the image features $\hat{I}$.
For this we leverage deformable cross-attention, originally proposed by \cite{li2022bevformer} and widely adopted in subsequent works.
In this operation, the current positions of the Gaussians $\mathcal{G}_\mu$ are projected onto the image feature maps using the camera parameters.
For each projected point, a set of image features around this point is sampled and used as keys and values for an attention operation.
This allows the Gaussians to extract rich, localized information from the scene in a computationally efficient manner.
For a detailed description of deformable attention, we refer readers to \cite{li2022bevformer}.

\begin{figure}
    \centering
    \resizebox{\linewidth}{!}{\includegraphics[page=3, trim=0cm 13.2cm 15.07cm 0cm, clip]{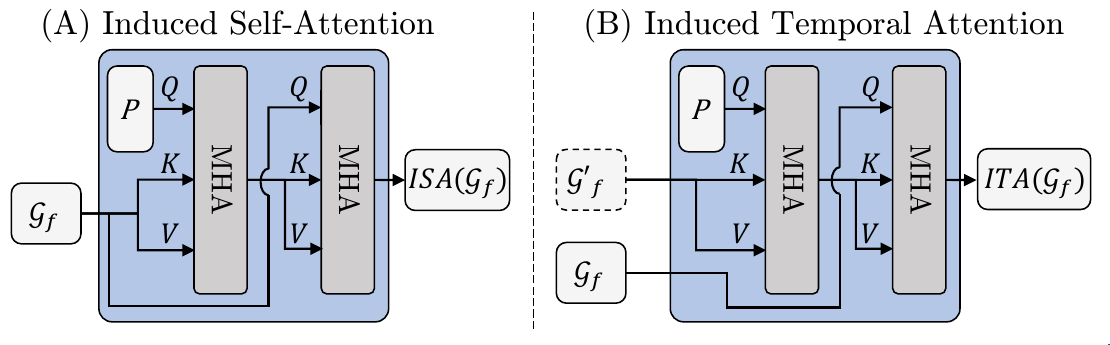}}
    \caption{
        \textbf{Illustration of the Induced Attention Modules.} 
        (A) Induced Self-Attention (ISA) 
        (B) Induced Temporal Attention (ITA)}
    \label{fig:induced_attn}
    \vspace{-4mm}
\end{figure}

\paragraph{Induced Self-Attention}
Computing interactions among Gaussians is a critical aspect of our approach.
However, the standard attention mechanism is characterized by quadratic time and memory complexity $\mathcal{O}(N^2)$.
This imposes severe constraints on the number of Gaussians that can be processed (as demonstrated in \cref{fig:ablation_induced}).
To address this limitation, we draw inspiration from the Set Transformer~\cite{lee2019set} and employ an attention operation called \emph{Induced Self-Attention} (ISA).
ISA reduces the memory complexity to approximately linearly scaling with respect to the number of Gaussians, allowing to employ a significantly larger number of Gaussians. 
The operation is schematically illustrated in \cref{fig:induced_attn} (A).
In essence, ISA replaces quadratic attention with two attention operations connected via a bottleneck.
Instead of allowing Gaussians to directly attend to one another, we introduce a set of $M$ trainable latent feature vectors, called inducing points $P \in \mathbb{R}^{M \times D}$ (with $M \ll N$).
These inducing points aggregate information from all Gaussians into a bottleneck representation $H$.
Subsequently, all Gaussians interact with the bottleneck features $H$ enabling indirect yet complete Gaussian-to-Gaussian interactions.
This transformation reduces the computational complexity to $\mathcal{O}(MN)$. 
The overall process is formally defined as:
\begin{align}
    &\ \mathrm{ISA}(\mathcal{G}_f) = \mathrm{MHA}(\mathcal{G}_f, H, H), \; \\
    &\ \text{where} \quad H = \mathrm{MHA}(P, \mathcal{G}_f, \mathcal{G}_f),
    \label{eq:induced_self}
\end{align}
where $\mathrm{MHA}(Q, K, V)$ denotes the standard multi-head attention operation, which includes a skip connection and a feed-forward layer.

\paragraph{Induced Temporal Attention}\label{sec:ita}
Building upon the concept of ISA, we introduce \emph{Induced Temporal Attention} (ITA) to enable efficient temporal information propagation across frames, as depicted in \cref{fig:induced_attn} (B).
The ITA operation leverages the structure of induced attention by letting the inducing points $P$ first attend to the Gaussian features from the previous frame to compute the bottleneck features $Z$.
Subsequently, Gaussian features from the current frame interact with the bottleneck features, facilitating temporal attention with linear complexity.
Formally, given $\mathcal{G}'_f$ as the final Gaussian features from the previous frame, the ITA operation is expressed as:
\begin{align}
    &\ \mathrm{ITA}(\mathcal{G}_f, \mathcal{G}'_f)= \mathrm{MHA}(\mathcal{G}_f, Z, Z), \; \\
    &\ \text{where} \quad Z = \mathrm{MHA}(P, \mathcal{G}'_f, \mathcal{G}'_f).
    \label{eq:induced_temp}
\end{align}

\paragraph{Gaussian Rectification}
Thus far, only the Gaussian features have been updated using cross-attention, self-attention and temporal attention.
In the \emph{Gaussian Rectification} step, these features are used to estimate an update to the Gaussian means for the next block. 
Unlike prior approaches such as GaussianFormer~\cite{huang2024gaussianformer} and GaussTR~\cite{jiang2024gausstr}, our method refrains from modifying Gaussian properties other than the means within the transformer itself.
Instead, we focus on gathering features that encapsulate all the necessary information.
The update is performed using an MLP, denoted as $\mathrm{Rect}(\mathcal{G}_f)$, which estimates a residual for each Gaussian and adds it to the current means to refine their position:
\begin{align}
    &\ \mathcal{G}^{b+1}_\mu = \mathcal{G}^{b}_\mu + \Delta \mathcal{G}^{b}_\mu, \; \\
    &\ \text{where} \quad \Delta \mathcal{G}^{b}_\mu = \mathrm{Rect}(\mathcal{G}^{b}_f).
\end{align}

\subsubsection{Gaussian Heads} \label{sec:gaussian_head}
After refining the Gaussian positions and features through $B$ blocks of the \emph{Gaussian Transformer}, the \emph{Gaussian Heads} estimate the remaining properties of the Gaussians, completing the scene representation.
This is accomplished using a set of individual MLP heads, each dedicated to predict a specific property from the Gaussian features.
To ensure valid outputs, a sigmoid activation $\sigma(\cdot)$ function is applied to estimate the opacity $o$, while the rotation coefficients $r$ are normalized to unit length.
The outputs are computed as follows:
 \begin{align}
    &\ o = \sigma \left(\mathrm{MLP}_o \left(\mathcal{G}_f \right) \right), \ s = \mathrm{MLP}_s(\mathcal{G}_f), \\ 
    &\ r = \mathrm{norm}(\mathrm{MLP}_r(\mathcal{G}_f)), \ c = \mathrm{MLP}_c(\mathcal{G}_f).
\end{align}
The outputs generated by the \emph{Gaussian Heads} define the final 3D scene representation, which can be used for downstream applications.

\subsection{Temporal Propagation}
As we show in \cref{fig:architecture} and \cref{fig:gauss_transformer}, the previous Gaussian features $\mathcal{G}'_f$ are derived by first correcting the previous final Gaussian means using ego-motion and estimated temporal flow to the current frame, computing the positional encoding and adding them to the previous final features.
The ITA module then fuses the temporal information in a recurrent manner.
The model always operates on inputs from a single time step while maintaining a memory of the last prediction to facilitate effective temporal fusion.

\subsection{Gaussian Splatting Supervision} \label{sec:gaussian_splatting}
Representing the scene with 3D Gaussians allows us to leverage the highly efficient Gaussian Splatting (GS)~\cite{kerbl20233d} method during training. 
For a detailed explanation of GS, we refer to the original work \cite{kerbl20233d}.
Essentially, the estimated Gaussian properties are used to project the 3D Gaussians back onto the input image views via the corresponding camera parameters.
This enables us to rasterize predicted 2D semantic logits and depth values.
Notably, unlike standard GS, which rasterizes RGB values, we rasterize semantic logits instead.
We generate 2D semantic and depth labels using pretrained models to serve as ground truth. 
The entire model is then trained using 2D rendering losses. 
This enables training in a weakly supervised manner directly in the 2D image space, eliminating the need for 3D labels or additional sensor data, such as LiDAR, entirely.
We optimize the MSE-Loss $\mathcal{L}_{depth}$ between rendered depth $\hat{D}$ and precomputed depth maps $D$ and the binary cross-entropy loss $\mathcal{L}_{seg}$ between rendered logits $\hat{S}$ and precomputed semantic segmentation maps $S$:
 \begin{align}
    \mathcal{L} = \mathcal{L}_{depth}(\hat{D}, D) + \mathcal{L}_{seg}(\hat{S}, S).
\end{align}

\subsection{Temporal Gaussian Splatting} \label{sec:temp_splatting}
Cameras in autonomous driving scenarios usually share a relatively small frustum overlap, which complicates learning of correct depth and scene geometry.
To address this limitation and increase viewpoint overlap between training cameras, we incorporate \emph{Temporal Gaussian Splatting}, a technique commonly employed in recent works~\cite{pan2023renderocc, boeder2024occflownet, gan2024gaussianocc}.
During training, for a specified temporal horizon $T$, we load the camera parameters and 2D ground truth labels of the $T$ previous and $T$ subsequent frames of the sequence.
We then use GS to rasterize our estimated 3D Gaussians into predicted depth and semantic maps for all temporal cameras and compute the same 2D rendering losses as for the current frame.
This strategy effectively widens the supervisory signal by incorporating data from neighboring temporal views.
While na\"ively applying \emph{Temporal Gaussian Splatting} can already significantly improve prediction performance, it overlooks the fact that many objects in the scene are in motion.
This motion introduces inconsistencies in supervisory signals across frames, as objects predicted in the current frame might have moved in previous/subsequent frames, so the 2D losses can induce erroneous signals.

\subsection{Temporal Module and Gaussian Flow} \label{sec:gaussian_flow}
To address temporal inconsistencies caused by the motion of dynamic objects during training, we introduce a \emph{Temporal Module} that learns to correct object motion.
This module reduces discrepancies across temporally adjacent frames by estimating 3D offsets for each Gaussian to align them with all frames within a defined temporal horizon.

The \emph{Temporal Module} takes the Gaussian features output by the \emph{Gaussian Transformer} as input and computes a 3D motion offset for each Gaussian relative to each temporal frame in the horizon.
To achieve this, we first define a set of learnable $D$-dimensional \textit{time tokens} $\Psi \in \mathbb{R}^{2T \times D}$, one token for each frame in the temporal horizon ($[-T, T]$).
For a given target time step, we create a copy of the Gaussians and append the corresponding time token to their features.
These temporally encoded features are then passed through an MLP to estimate a 3D motion offset vector $\overrightarrow{v} (t) \in \mathbb{R}^3$:
 \begin{align}
    \overrightarrow{v} (t) = \mathrm{MLP}_v  \left(\mathcal{G}_f \oplus  \Psi(t) \right).
\end{align}
The estimated offsets are added to the copied Gaussian positions, effectively translating the Gaussians to their updated locations in the target time step.
GS is subsequently applied in the usual manner, using the translated Gaussians for the corresponding temporal frame.
The \emph{Temporal Module} is trained jointly with the rest of the model.
Importantly, it does not require additional losses or ground truth motion data.
Dynamic objects are correctly rendered in temporally adjacent frames only when their motion is accurately estimated, allowing the module to learn implicitly through the existing rendering losses.
The benefits of this dynamic object handling are evaluated in \cref{sec:ablation_temp_module}, demonstrating its impact on scene consistency and model performance.

\subsection{Voxelization} \label{sec:voxelize}
The final output of our model consists solely of 3D Gaussian distributions (see \cref{fig:teaser}), with no voxelization or dense operations used during the model’s architecture or training process.
However, these Gaussians can be efficiently voxelized as a post-processing step to facilitate direct comparison with prior work on occupancy estimation benchmarks.
For each voxel center with position $p_i=(x,y,z)$ on a defined voxel grid, the estimated 3D Gaussian distributions are queried and their opacity and semantic logits are accumulated at each point.
Each voxel is assigned a label corresponding to the highest accumulated semantic logits.
Voxels with an accumulated opacity below a defined threshold are classified as free.
Further details about the voxelization are provided in the supplementary material.

\section{Experiments} \label{sec:experiments}
To evaluate the performance of GaussianFlowOcc, we compare our approach with recent state-of-the-art methods on occupancy estimation.
In addition, an extensive ablation study is performed to study the impact of each module. 

\subsection{Dataset and Metrics}
We conduct experiments on the Occ3D-nuScenes dataset~\cite{tian2023occ3d}, a widely used benchmark for occupancy estimation based on the nuScenes dataset~\cite{caesar2020nuscenes}.
The dataset defines a voxel grid spanning $[-40m, 40m]$ along the X and Y axes and $[-1m, 5.4m]$ along the Z axis, with a voxel size of $0.4m$.
Each voxel is assigned one of $17$ semantic labels from nuScenes Lidarseg~\cite{fong2021panoptic} or classified as free space.
Performance on this benchmark is evaluated using the Intersection over Union (IoU) score for each of the $17$ semantic categories, aggregated to the mean IoU (mIoU).
Additionally, following recent trends in the literature, we also report the RayIoU metric~\cite{liu2024fully}, a ray-based metric focused on rewarding good 3D completion while penalizing thick surface predictions.
Finally, we provide information on inference speed (FPS) on a single A100 GPU.

\subsection{Implementation Details}
We utilize a ResNet-50~\cite{he2016deep} backbone for feature extraction with an image resolution of $256\times704$ and set the number of Gaussians to $N=10000$.
The \emph{Gaussian Transformer} consists of three blocks, each with induced attention layers employing $M=500$ inducing points.
The \emph{Gaussian Heads} are implemented as a single linear layer, while all layers in the model share a latent dimension of $256$.
The \emph{Temporal Module} consists of three linear layers and operates with a temporal horizon $T=6$ during training.
The model is trained for $18$ epochs using four A100 GPUs.
The source code is available at \url{https://github.com/boschresearch/GaussianFlowOcc}.

\subsection{3D Occupancy Prediction Results}

\begin{table*}[ht]
    \begin{center}
        \caption{
            \textbf{Occupancy estimation performance on the Occ3D-nuScenes validation dataset.}
            The \emph{Training Labels} column specifies the method used for pseudo label generation.
            Performance is reported in terms of IoU (\%).
            The best-performing result in each column is highlighted in \textbf{bold}, while the second-best is shown in \textit{italic}. 
            A dash (-) indicates that no results were provided by the original work.}
        \label{table:main}
        \resizebox{\textwidth}{!}{%
            \begin{tabular}{lll|cc|cccc|c}
                \hline
                \noalign{\smallskip}
                 Method & Backbone & Training Labels & mIoU & IoU & RayIoU & RayIoU@1 & RayIoU@2 & RayIoU@4 & FPS\\
                \noalign{\smallskip}
                \hline
                \noalign{\smallskip}
                SelfOcc \cite{huang2023selfocc} & ResNet-50 & OpenSeeD & 10.54 & 45.01 & - & - & - & -  & 1.15 \\
                OccNeRF \cite{zhang2023occnerf} & ResNet-101 & GroundedSAM & 10.81 & 22.81 & - & - & - & - & 1.27 \\
                GaussianOcc \cite{gan2024gaussianocc} & ResNet-101 & GroundedSAM & 11.26 & - & 11.85 & \textit{8.69} & 11.90 & 14.95 & \textit{5.57} \\
                GaussTR \cite{jiang2024gausstr} & $2 \times$ViT & CLIP + Metric3D + GroundedSAM & \textit{13.26} & \textit{45.19} & - & - & - & - & 0.20 \\
                \noalign{\smallskip}
                \hline
                \noalign{\smallskip}
                GaussianFlowOcc (Ours) & ResNet-50 & GroundedSAM + Metric3D & \textbf{17.08} & \textbf{46.91} & \textbf{16.47} & \textbf{11.81} & \textbf{16.58} & \textbf{20.98} & \textbf{10.2} \\
                \noalign{\smallskip}
                \hline
            \end{tabular}
            \addtolength{\tabcolsep}{2pt}
        }
    \end{center}
    \vspace{-4mm}
\end{table*}

We evaluate our proposed method against state-of-the-art approaches for weakly supervised 3D occupancy prediction on the Occ3D-nuScenes dataset~\cite{tian2023occ3d}, with the results summarized in \cref{table:main}.
Each of the compared methods relies on a pretrained semantic segmentation model to generate 2D semantic labels for training.
Our approach significantly outperforms all existing models, achieving at least a $51\%$ mIoU improvement over methods that rely on RGB reconstruction and pseudo semantic labels.
Notably, we surpass the recent GaussTR, which also uses Metric3D for depth supervision, by $29\%$ mIoU, improving from $13.26$ to $17.08$.
Furthermore, our model achieves this superior performance with a rather lightweight ResNet-50 backbone, while GaussTR uses CLIP and Metric3D as backbones during inference, which is significantly more costly.
A similar trend is observed in the RayIoU metric, where our method improves performance by approximately $39\%$ compared to GaussianOcc.
Beyond accuracy, our model also offers a substantial efficiency advantage.
Due to its lightweight architecture, it achieves inference speeds up to almost $9\times$ faster than previous voxel-based methods.
As mentioned above, especially GaussTR remains computationally expensive due to its reliance on multiple ViT-L backbones; GaussianFlowOcc is $50 \times$ faster.

We attribute these performance gains primarily to two key innovations: 
Firstly, our \emph{Temporal Module} significantly improves temporal consistency, yielding large performance increases as we show in \cref{sec:ablation_temp_module}. 
Secondly, the \emph{Gaussian Transformer} allows for a much higher number of Gaussians to be processed efficiently, due to approximately linearly scaling induced attention modules.
A detailed per-class performance breakdown is provided in the supplement.
Additionally, qualitative results presented in \cref{fig:teaser} and in the supplement highlight our model’s strong 3D scene completion abilities.
In particular, our approach excels at representing small, thin, and flat objects (e.g., traffic signs, poles, pedestrians), a major advantage of using 3D Gaussians.
In contrast, voxel-based methods are inherently constrained by a grid resolution, limiting their ability to capture fine details.
\subsection{Ablation Study} \label{sec:exp3}

\begin{table*}[t] 
    \begin{minipage}[t]{0.3\textwidth}
        \centering
        \caption{
            \textbf{Ablation on self and temporal attention.}
            Performance changes when omitting self and/or temporal attention within the \emph{Gaussian Transformer}.}
        \label{table:ablation_attention}
        \resizebox{\textwidth}{!}{
        \begin{tabular}{cc|c}
            \hline
            \thead{Induced \\ Self-Attention} & \thead{Induced \\ Temporal Attention} & mIoU \\
            \hline
            & & 13.81 \\
             \checkmark & & 14.60 \\
              & \checkmark & 14.47 \\
            \checkmark & \checkmark &  \textbf{17.08} \\
            \hline
        \end{tabular}
        }
    \end{minipage}
    \hfill
    \begin{minipage}[t]{0.3\textwidth}
        \centering
         \caption{
            \textbf{Ablation study of the proposed \emph{Temporal Module}.}
            Performance comparison when compensating for object motion during training versus ignoring temporal inconsistencies during \emph{Temporal Gaussian Splatting}.
            }
        \label{table:ablation_temp_module}
        \resizebox{\textwidth}{!}{\begin{tabular}{c|cc}
            \hline
            Temporal Module & mIoU & RayIoU \\
            \hline
             & 14.18 & 14.46 \\
            \checkmark & \textbf{17.08} &  \textbf{16.47} \\
            \hline
        \end{tabular}}
    \end{minipage}
    \hfill
    \begin{minipage}[t]{0.3\textwidth}
        \centering
         \caption{
            \textbf{Ablation on the Gaussian parameters.}
            Performance values when only using and learning a subset of the Gaussian parameters.
            }
        \label{table:ablation_params}
        \resizebox{\textwidth}{!}{\begin{tabular}{ccc|c}
            \hline
             Opacity & Scale & Rotation & mIoU \\
            \hline
             & & & 9.48 \\
             \checkmark & & & 13.12 \\
             \checkmark & \checkmark & & 14.98 \\
             \checkmark & \checkmark & \checkmark & \textbf{17.08}\\
            \hline
        \end{tabular}}
    \end{minipage}
\end{table*}

\subsubsection{Induced Attention vs. Full Attention} \label{sec:ablation_induced}
\begin{figure}
    \centering
	\resizebox{\linewidth}{!}{\includegraphics[page=7, trim=0cm 9.89cm 13.23cm 0cm, clip]{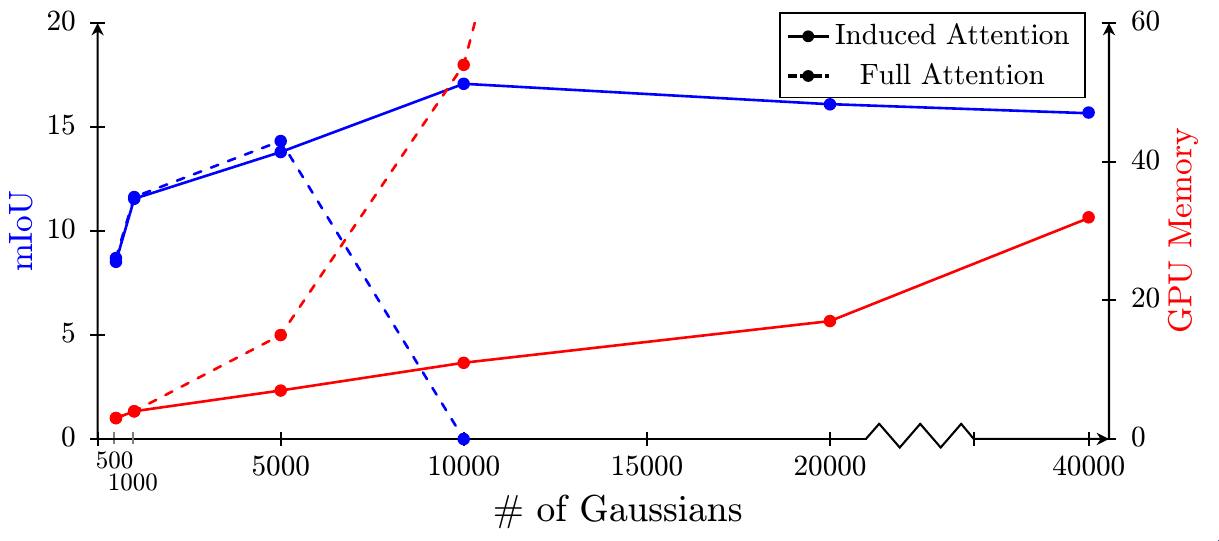}}
	\caption{
		\textbf{Comparison in mIoU and GPU memory consumption between Induced and Full Attention.}}
	\label{fig:ablation_induced}
	\vspace{-4mm}
\end{figure}

To verify the effectiveness of the proposed induced attention mechanism, we compare it to a standard full attention mechanism.
The results are shown in \cref{fig:ablation_induced}.
The plot shows both mIoU score and memory consumption for a single sample during training for induced attention (with fixed $M=500$) and standard attention.
As can be seen, with lower numbers of Gaussians, the induced attention and full attention exhibit comparable performance and memory usage, as the quadratic complexity of the full attention is not yet significant.
Increasing the number of Gaussians to $N=5000$ already requires double the amount of memory for full attention, while not providing any significant performance increase.
Using $10000$ Gaussians demands $50\mathrm{GB}$ of GPU memory, forcing us to reduce the batch size to $1$ on our hardware.
Training a model this way did not lead to convergence.
In contrast, using induced attention, the memory requirement scales linearly, ultimately enabling better performance.
The increase in mIoU achieved by the Induced Attention model when increasing the number of Gaussians from $5000$ to $10000$ indicates that a higher number of Gaussians is necessary to represent the complexity of driving scenes.
We find that increasing $N$ beyond $10000$ did not improve performance.

\subsubsection{Self-Attention and Temporal Attention}
We also show the necessity of the self-attention and temporal attention mechanisms by training a model without each of these modules.
The results are displayed in \cref{table:ablation_attention}, indicating that both modules increase performance significantly.
Note that GICA is always used, as no scene information would be used otherwise.

\subsubsection{Temporal Module} \label{sec:ablation_temp_module}
We demonstrate the effectiveness of the proposed \emph{Temporal Module} by training a model without it while keeping all other components unchanged.
As visible in \cref{table:ablation_temp_module}, using the \emph{Temporal Module} leads to a relative improvement in mIoU of $20\%$ and in RayIoU of $14\%$, which emphasizes the importance of object motion compensation during training.
A demonstration of rasterized views of temporally adjacent frames with and without applied estimated flow is provided in the supplement and in \cref{fig:teaser}.

\subsubsection{Gaussian Parameters}
In contrast to previous works \cite{gan2024gaussianocc, sun2024gsrender}, we find that allowing the model to freely adjust all Gaussian parameters yields the best results. 
To demonstrate this, we conduct an experiment (\cref{table:ablation_params}) where we disable the \emph{Gaussian Heads} for certain properties and fix their values, preventing them from being learned.
We set the opacity to $1$, the scale to $0.3$ and the rotations to unit quaternions.
When only estimating the mean and fixing all other properties, only a performance of $9.48$ mIoU can be achieved, showing the importance of the other properties.
When allowing the model to influence the Gaussians opacity, a large improvement to $13.12$ mIoU is reached, outperforming most previous methods.
Adding the other properties further increases performance, with the maximum score achieved when using all properties.
As mentioned above, the positive effect of using rotation and scale is especially noticeable for thin or flat structures.

\subsubsection{Time Horizon}
We ablate the \emph{Temporal Gaussian Splatting} horizon $T$ in the range of $\{0, 2, ..., 10\}$ to demonstrate how temporal rendering leads to better supervision, shown in \cref{table:ablation_horizon}.
With horizons greater than 8, the model diverges, which is likely due to large discrepancy between the time steps, making it increasingly harder to estimate (especially non-linear) motion. 
Additionally, some camera poses are just too distant from the current time step.

\begin{table}
\vspace{-4mm}
    \begin{center}
        \caption{\textbf{Ablation on different horizons $T$. }
            The best performance is achieved with $T=6$. 
            The model diverges with $T\geq10$.}
        \label{table:ablation_horizon}
     \resizebox{\columnwidth}{!}{
        \begin{tabular}{c|cccccc}
            \hline
            Time Horizon & 0 & 2 & 4 & \textbf{6} & 8 & 10\\
            \hline
            mIoU & 12.67 & 14.17 & 14.71 & \textbf{17.08} & 14.82 & N/A \\
            \hline
        \end{tabular}
        }
    \end{center}
    \vspace{-4mm}
\end{table}

\subsubsection{Pseudo Depth Labels}
To analyze the impact of pseudo depth labels generated by Metric3D, we train a model using only pseudo semantics.
The results of this experiment are presented in \cref{table:ablation_pseudodepth}.
Even without pseudo depth supervision, our model outperforms all prior works, demonstrating its strong capability.
Notably, other approaches \cite{zhang2023occnerf, huang2023selfocc} incorporate additional photometric losses to enhance geometric consistency, which we did not find to be beneficial.
Interestingly, GaussTR~\cite{jiang2024gausstr} fails to converge without depth labels, further highlighting the robustness of our approach.
This suggests that \emph{Temporal Gaussian Splatting} alone provides sufficient 3D cues for geometry estimation by leveraging semantic maps from multiple cameras over an extended time horizon.

\begin{table}
    \begin{center}
    \small
        \caption{\textbf{Ablation on pseudo depth labels.}
        Depth labels substantially improve performance.}
        \label{table:ablation_pseudodepth}
        \begin{tabular}{c|cc|c}
            \hline
            Pseudo Depth & mIoU & IoU & RayIoU \\
            \hline
             & 14.07 & 34.12 & 12.75 \\
            \checkmark & \textbf{17.08} & \textbf{46.91} & \textbf{16.47}\\
            \hline
        \end{tabular}
        
    \end{center}
    \vspace{-4mm}
\end{table}


\section{Conclusion and Future Work} \label{sec:conclusion}
In this work, we introduced a novel, linearly scaling Transformer architecture for efficiently modeling driving scenes using 3D Gaussian distributions, trained with \emph{Temporal Gaussian Splatting} and explicit object motion modeling.
Our approach significantly outperforms previous weakly supervised methods on the Occ3D-nuScenes benchmark while maintaining substantially faster inference speeds.
Despite these advances, there are still areas for improvement.
We found that self-supervised learning (SSL) with RGB rendering did not provide benefits, suggesting that further investigation is needed in this direction.
Moreover, a more sophisticated approach to dynamic object modeling could further improve temporal consistency.
Incorporating physically informed motion priors, leveraging temporal and Gaussian dependencies as well as adapting more properties than just the mean to better represent deformable objects could enhance temporal modeling.
For future work, the sparse and efficient attention mechanism proposed in this work could be applied to other downstream tasks, such as object detection, lane segmentation, and planning. 
This opens up exciting possibilities for using our model in autonomous driving and robotics applications.

{
    \small
    \bibliographystyle{ieeenat_fullname}
    \bibliography{main}
}
 
\clearpage
\setcounter{page}{1}
\maketitlesupplementary
\appendix
\renewcommand\thefigure{\thesection.\arabic{figure}}
\setcounter{figure}{0}

\renewcommand\thetable{\thesection.\arabic{table}}
\setcounter{table}{0}

\section{Additional Quantitative Results}

\subsection{Per-Class IoU}
For completeness, we provide per-class IoU scores for all evaluated methods on the Occ3D-nuScenes dataset in \cref{table:main_complete}.
Our proposed GaussianFlowOcc consistently outperforms all prior methods trained with pseudo labels, both with and without depth supervision.
The most significant performance gains can be observed in small object classes, such as motorcycles and traffic cones, but also in other categories.
Notably, our method achieves the best or second best score across nearly all categories. 
These results further highlight the effectiveness of our Gaussian-based scene representation and temporal modeling, particularly in handling fine-grained object details that are often challenging for voxel-based approaches.

\begin{table*}
    \begin{center}
        \caption{
            \textbf{Occupancy estimation performance on the Occ3D-nuScenes validation set.}
            The \textit{Mode} indicates the source of the 2D labels.
            $C$ refers to camera images, $D$ refers to usage of pseudo depth.
            Best performing per column and \textit{Mode} section in \textbf{bold}, second best in \textit{italics}.
            Methods that use pseudo labels ignore the \textit{others} and \textit{other flat} classes.}
        \label{table:main_complete}
        
        \resizebox{\textwidth}{!}{%
            \addtolength{\tabcolsep}{2pt}
            \begin{tabular}{ll|c|ccccccccccccccccc}
                \hline
                \noalign{\smallskip}
                 Method & \rotatebox{90}{Mode} & mIoU & \rotatebox{90}{others} & \rotatebox{90}{barrier} & \rotatebox{90}{bicycle} & \rotatebox{90}{bus} & \rotatebox{90}{car} & \rotatebox{90}{cons. vehicle} & \rotatebox{90}{motorcycle} & \rotatebox{90}{pedestrian} & \rotatebox{90}{traffic cone} & \rotatebox{90}{trailer} & \rotatebox{90}{truck} & \rotatebox{90}{driv. surf.} & \rotatebox{90}{other flat} & \rotatebox{90}{sidewalk} & \rotatebox{90}{terrain} & \rotatebox{90}{manmade} & \rotatebox{90}{vegetation}\\

                \noalign{\smallskip}
                \hline
                \noalign{\smallskip}
                
                SelfOcc \cite{huang2023selfocc} & C & 10.54 & - & 0.15 & 0.66 & 5.46 & \textit{12.54} & 0.00 & 0.80 & 2.10 & 0.00 & 0.00 & 8.25 & \textbf{55.49} & - & \textit{26.30} & \textit{26.54} & \textit{14.22} & 5.60 \\
                OccNeRF \cite{zhang2023occnerf} & C & 10.81 & - & 0.83 & 0.82 & 5.13 & 12.49 & \textit{3.50} & 0.23 & 3.10 & 1.84 & 0.52 & 3.90 & 52.62 & - & 20.81 & 24.75 & \textbf{18.45} & \textbf{13.19} \\
                GaussianOcc \cite{zhang2023occnerf} & C & \textit{11.26} & - & \textit{1.79} & \textit{5.82} & \textbf{14.58} & \textbf{13.55} & 1.30 & \textit{2.82} & \textbf{7.95} & \textbf{9.76} & \textit{0.56} & \textit{9.61} & 44.59 & - & 20.10 & 17.58 & 8.61 & \textit{10.29} \\
                
               \textit{GaussianFlow (Ours)} & C & \textbf{14.07} & - & \textbf{6.27} & \textbf{8.54} & \textit{13.36} & 12.38 & \textbf{4.92} & \textit{10.05} & \textit{6.84} & \textit{8.75} & \textbf{1.12} & \textbf{10.43} & \textit{53.40} & - & \textbf{26.44} & \textbf{28.89} & 10.39 & 9.33 \\
               
                \noalign{\smallskip}
                \hline
                \noalign{\smallskip}
                
                GaussTR \cite{jiang2024gausstr} & C+D & \textit{13.26} & - & \textit{2.09} & \textit{5.22} & \textit{14.07} & \textbf{20.43} & \textbf{5.70} & \textit{7.08} & \textit{5.12} & \textit{3.93} & \textit{0.92} & \textbf{13.36} & \textit{39.44} & - & \textit{15.68} & \textit{22.89} & \textbf{21.17} & \textbf{21.87} \\
                \textit{GaussianFlow (Ours)} & C+D & \textbf{17.08} & - & \textbf{6.75} & \textbf{9.68} & \textbf{18.98} & \textit{17.15} & \textit{4.19} & \textbf{11.78} & \textbf{9.27} & \textbf{10.30} & \textbf{1.83} & \textit{12.33} & \textbf{61.03} & - & \textbf{31.17} & \textbf{34.78} & \textit{14.66} & \textit{12.40} \\
                \noalign{\smallskip}
                \hline
            \end{tabular}
            \addtolength{\tabcolsep}{2pt}
        }
    \end{center}
\end{table*}


\subsection{Depth Estimation}
To further demonstrate the 3D geometric capabilities of GaussianFlowOcc, we report depth estimation results on the nuScenes dataset in \cref{table:depth}. Our method achieves competitive performance compared to other state-of-the-art self-supervised and weakly supervised approaches on this task.
We observe that methods utilizing photometric losses~\cite{huang2023selfocc, zhang2023occnerf, gan2024gaussianocc} tend to produce slightly more accurate camera depth predictions than our approach. However, it is important to note that while our method may yield marginally lower depth estimation accuracy, it substantially outperforms these methods in 3D occupancy estimation, underscoring the strength of our model in comprehensive 3D scene understanding.

\begin{table*}
    \begin{center}
        \caption{\textbf{Depth estimation results on nuScenes.} 
        The best result is highlighted in \textbf{bold}, second best in \textit{italics}.}
        \label{table:depth}
     \resizebox{\textwidth}{!}{
        \begin{tabular}{l|ccccccc}
            \hline
            \noalign{\smallskip}
            Method & Abs Rel $\downarrow$ & Sq Rel $\downarrow$ & RMSE $\downarrow$ & RMSE log $\downarrow$ & $\delta < 1.25 \uparrow$ & $\delta < 1.25^2 \uparrow$ & $\delta<1.25^3 \uparrow$\\
            \noalign{\smallskip}
            \hline
            \noalign{\smallskip}
            SurroundDepth \cite{wei2023surrounddepth} & 0.280 & 4.401 & 7.467 & 0.364 & 0.661 & 0.844 & 0.917 \\
            SimpleOcc \cite{gan2023simple} & 0.224 & 3.383 & 7.165 & 0.333 & \textit{0.753} & \textit{0.877} & 0.930 \\
            OccNeRF \cite{zhang2023occnerf} & \textit{0.202} & 2.883 & \textit{6.697} & 0.319 & \textbf{0.768} & \textbf{0.882} & \textit{0.931} \\
            SelfOcc \cite{huang2023selfocc} & 0.215 & 2.743 & 6.706 & \textit{0.316} & \textit{0.753} & 0.875 & \textbf{0.932} \\
            GaussianOcc \cite{zhang2023occnerf} & \textbf{0.197} & \textbf{1.846} & 6.733 & \textbf{0.312} & 0.746 & 0.873 & \textit{0.931} \\
            \noalign{\smallskip}
            \hline
            \noalign{\smallskip}
            GaussianFlowOcc (Ours) & 0.278 & \textit{2.522} & \textbf{5.232} & 0.389 & 0.677 & 0.826 & 0.898\\
            \noalign{\smallskip}
            \hline
        \end{tabular}
        }
    \end{center}
\end{table*}

\subsection{Training with 3D Voxel Labels}\label{sec:3d_labels}
Our main contributions, namely the Gaussian representation and the dynamic object compensation during temporal Gaussian Splatting, specifically target the weakly supervised setting for occupancy estimation.
Nonetheless, our model can be trained using 3D voxel labels similar to previous methods \cite{li2023fb, li2022bevformer, zhang2023occformer, tian2023occ3d}.
Specifically, we apply a differentiable voxelization to the final Gaussian predictions (see \cref{sec:voxelize}), enabling the computation of a cross-entropy loss with ground-truth voxel labels for supervised training.
Note that for this experiment, the \textit{Temporal Module} and Gaussian Splatting supervision are not used.
In \cref{table:main_3d}, we present a comparison on the Occ3D-nuScenes dataset between our method and established fully supervised approaches, following the recommended protocol of not using the camera visibility mask~\cite{tian2023occ3d}.
As shown, our model outperforms prior methods such as BEVFormer~\cite{li2022bevformer} and OccFormer~\cite{zhang2023occformer}, while achieving competitive performance with larger models specifically designed for full supervision, including FB-Occ~\cite{li2023fb} and CTF-Occ~\cite{tian2023occ3d}.

\begin{table*}
    \begin{center}
        \caption{
            \textbf{Occupancy estimation performance on the Occ3D-nuScenes validation set using 3D voxel labels for training.}
            The best result per column is highlighted in \textbf{bold}, second best in \textit{italics}.
        }
        \label{table:main_3d}
        
        \resizebox{\textwidth}{!}{%
            \addtolength{\tabcolsep}{2pt}
            \begin{tabular}{ll|c|ccccccccccccccccc}
                \hline
                \noalign{\smallskip}
                 Method & \rotatebox{90}{Backbone} & mIoU & \rotatebox{90}{others} & \rotatebox{90}{barrier} & \rotatebox{90}{bicycle} & \rotatebox{90}{bus} & \rotatebox{90}{car} & \rotatebox{90}{cons. vehicle} & \rotatebox{90}{motorcycle} & \rotatebox{90}{pedestrian} & \rotatebox{90}{traffic cone} & \rotatebox{90}{trailer} & \rotatebox{90}{truck} & \rotatebox{90}{driv. surf.} & \rotatebox{90}{other flat} & \rotatebox{90}{sidewalk} & \rotatebox{90}{terrain} & \rotatebox{90}{manmade} & \rotatebox{90}{vegetation}\\
                \noalign{\smallskip}
                \hline
                \noalign{\smallskip}
                BEVFormer \cite{li2022bevformer} & ResNet-101 & 23.67 & 5.0  & 38.8 & 10.0  & 34.4 & \textit{41.1} & 13.2 & 16.5 & 18.2 & 17.8 & \textit{18.7} & \textit{27.7} & 49.0  & 27.7 & 29.1 & 25.4 & 15.4 & 14.5 \\
                OccFormer \cite{zhang2023occformer} & ResNet-101 & 21.93 & 5.9 & 30.3 & 12.3 & 34.4 & 39.2 & 14.4 & 16.4 & 17.2 &  9.3 & 13.9 & 26.4 & 51.0  & 31.0  & 34.7 & 22.7 &  6.8 &  7.0 \\
                FB-Occ \cite{li2023fb} & ResNet-50 & \textit{27.09} & 0.0  & \textbf{40.9} & \textbf{21.2} & \textbf{39.2} & 40.8 & \textbf{20.6} & \textit{23.8} & \textbf{23.6} & \textbf{25.0}  & 16.6 & 26.4 & \textit{59.4} & 27.6 & 31.4 & 29.0  & 16.7 & \textbf{18.4} \\
                CTF-Occ \cite{tian2023occ3d} & ResNet-101 & \textbf{28.53} & \textbf{8.1} & \textit{39.3} & \textit{20.6} & \textit{38.3} & \textbf{42.2} & \textit{16.9} & \textbf{24.5} & \textit{22.7} & \textit{21.0} & \textbf{23.0} & \textbf{31.1} & 53.3 & \textit{33.8} & \textit{38.0} & \textit{33.2} & \textbf{20.8} & \textit{18.0} \\
                               
                \noalign{\smallskip}
                \hline
                \noalign{\smallskip}

                GaussianFlowOcc (Ours) & ResNet-50 & 25.28 & \textit{7.6} & 25.4 & 13.4 & 26.0 & 25.3 & 15.3 & 14.2 & 12.7 & 10.7 & 18.2 & 20.9 & \textbf{76.2} & \textbf{37.4} & \textbf{48.2} & \textbf{47.8} & \textit{17.3} & 13.2 \\
                \noalign{\smallskip}
                \hline
            \end{tabular}
            \addtolength{\tabcolsep}{2pt}
        }
    \end{center}
\end{table*}

\section{Qualitative Results}

\subsection{Predicted Occupancy}
\begin{figure*}
    \centering
    \includegraphics[page=4, trim=0cm 5.47cm 11.11cm 0cm, clip, width=1\textwidth]{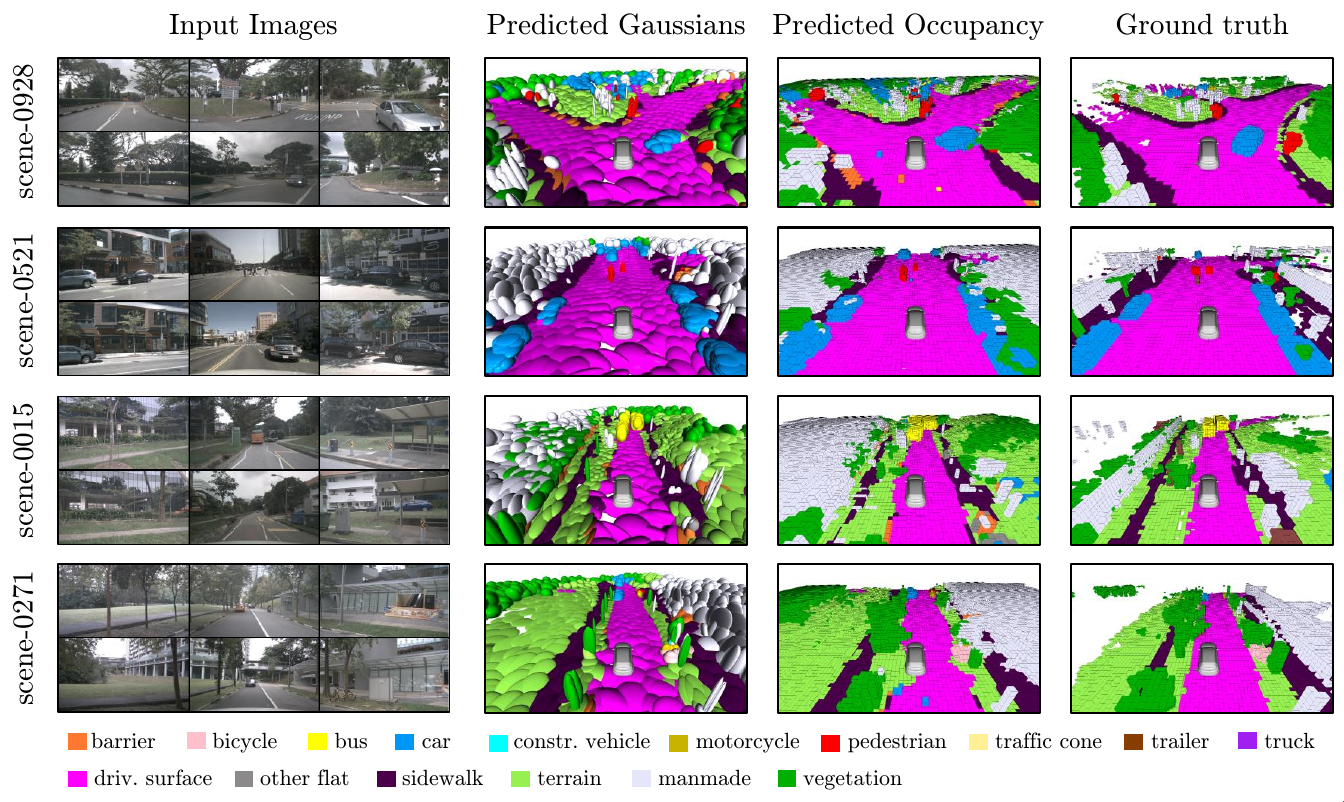}
    \caption{
        \textbf{Qualitative results on the Occ3D-nuScenes dataset.}
        We show the estimated 3D Gaussians, how the predictions look when voxelized, and the ground truth occupancy.
        Best viewed when zoomed in.
    }
    \label{fig:qualitative1}
\end{figure*}

In \cref{fig:qualitative1}, we present qualitative examples of GaussianFlowOcc's predicted occupancy compared to the ground truth, visualized from a third-person perspective.
These examples demonstrate our model's ability to precisely reconstruct the 3D scene geometry using 3D Gaussians, despite the absence of explicit 3D supervision.
The Gaussians align naturally to capture the shapes and details of objects, effectively representing scene structures without requiring per-scene optimization.
This highlights the flexibility and efficiency of our approach in modeling complex environments.
Additionally, we provide a set of videos in our GitHub repository (see \cref{sec:source_code}) showcasing inference results on full validation scenes:
\begin{itemize}
    \item \textit{scene-0346.mp4} and \textit{scene-0790.mp4} visualize the predicted Gaussians from three perspectives: input camera view, third-person view, and Bird's-Eye View.
    \item \textit{scene-0521\_gt.mp4} and \textit{scene-0925\_gt.mp4} compare our model’s predictions to the ground truth occupancy, both from a third-person perspective.
\end{itemize}

These results clearly demonstrate the model’s ability to accurately represent flat surfaces (e.g., streets and walls), thin objects (e.g., poles), and small objects (e.g., traffic cones).
Unlike voxel-based methods, which are constrained by coarse resolution, our approach captures continuous geometric details with far greater fidelity, enabling a more precise and realistic 3D scene understanding.

\subsection{Rendered Depth and Semantics}
In \cref{fig:qual_render}, we present examples of rendered depth and semantics obtained by applying Gaussian Splatting to the 3D Gaussians estimated from our trained model.
For comparison, we also show the pseudo labels used during training, which are generated by GroundedSAM~\cite{ren2024grounded} for semantics and Metric3D~\cite{yin2023metric3d} for depth.
The rendered outputs demonstrate a high level of detail, with most objects accurately segmented.
Notably, even thin structures such as trees and pedestrians are well-preserved, showcasing the model's ability to preserve fine geometric details; something voxel-based approaches often struggle with due to grid resolution constraints.

In \cref{fig:qual_render_temporal}, we further illustrate the effect of our \emph{Temporal Gaussian Splatting} by comparing predicted semantic segmentation maps across consecutive frames:
The first row shows the rendered semantics for the current input frame.
The second row depicts the segmentation map generated by rendering into the next frame and using the 3D flow estimated by the \emph{Temporal Module} to compensate the object motion.
The third row shows the same next-frame projection but without dynamic object handling — simulating the approach taken by previous works.
The last row provides the pseudo-semantic labels of the next frame for comparison.
The comparison clearly reveals the importance of dynamic object modeling.
Without applying the 3D flow (Row 3), the rendered semantics fail to align with the dynamic objects in the next frame, as these objects have already moved.
This misalignment would introduce incorrect supervisory signals if used to compute the loss.
In contrast, when applying the \emph{Temporal Module}’s flow estimation (Row 2), the rendered semantics better match the pseudo labels of the next frame.
This demonstrates that our \emph{Temporal Module} has successfully learned to approximate object motion, improving the alignment of dynamic objects across frames and ensuring more accurate supervision during training.
\afterpage{
\clearpage
    \begin{figure*}
      \centering
          \includegraphics[page=5, trim=0cm 4.89cm 7.29cm 0cm, clip, width=1\textwidth]{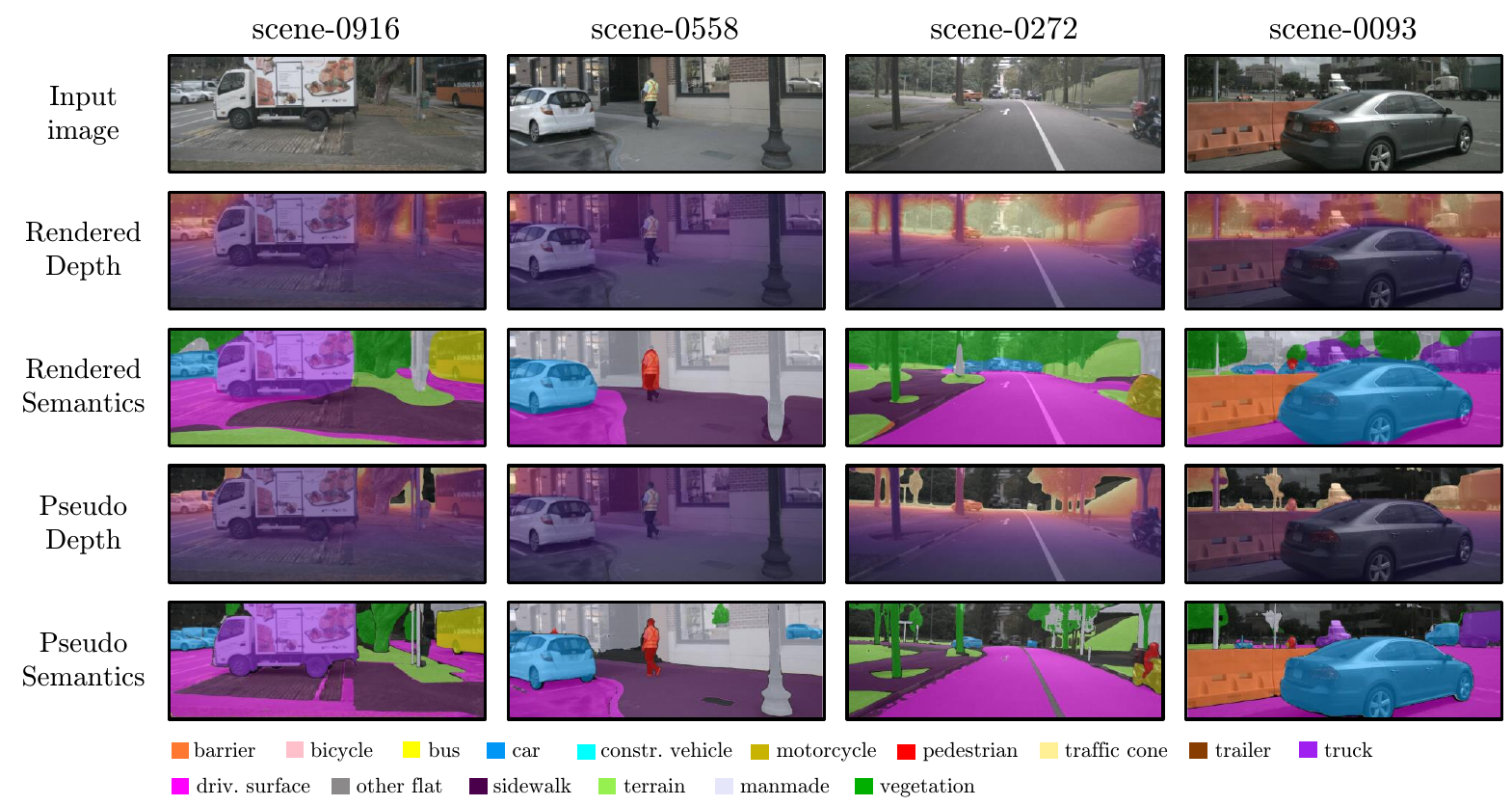}
      \caption{
      \textbf{Qualitative results on the Occ3D-nuScenes validation set.}
      Each column shows an \textit{Input Image}, \textit{Rendered Depth} and \textit{Rendered Semantics} generated by using GS to render predicted Gaussians into input cameras (as is done during training).
      We additionally show the pseudo labels used during training for the relevant sample.}
    \label{fig:qual_render}
    \end{figure*}
    
    \begin{figure*}
      \centering
          \includegraphics[page=6, trim=0cm 4.99cm 6.11cm 0cm, clip, width=.99\textwidth]{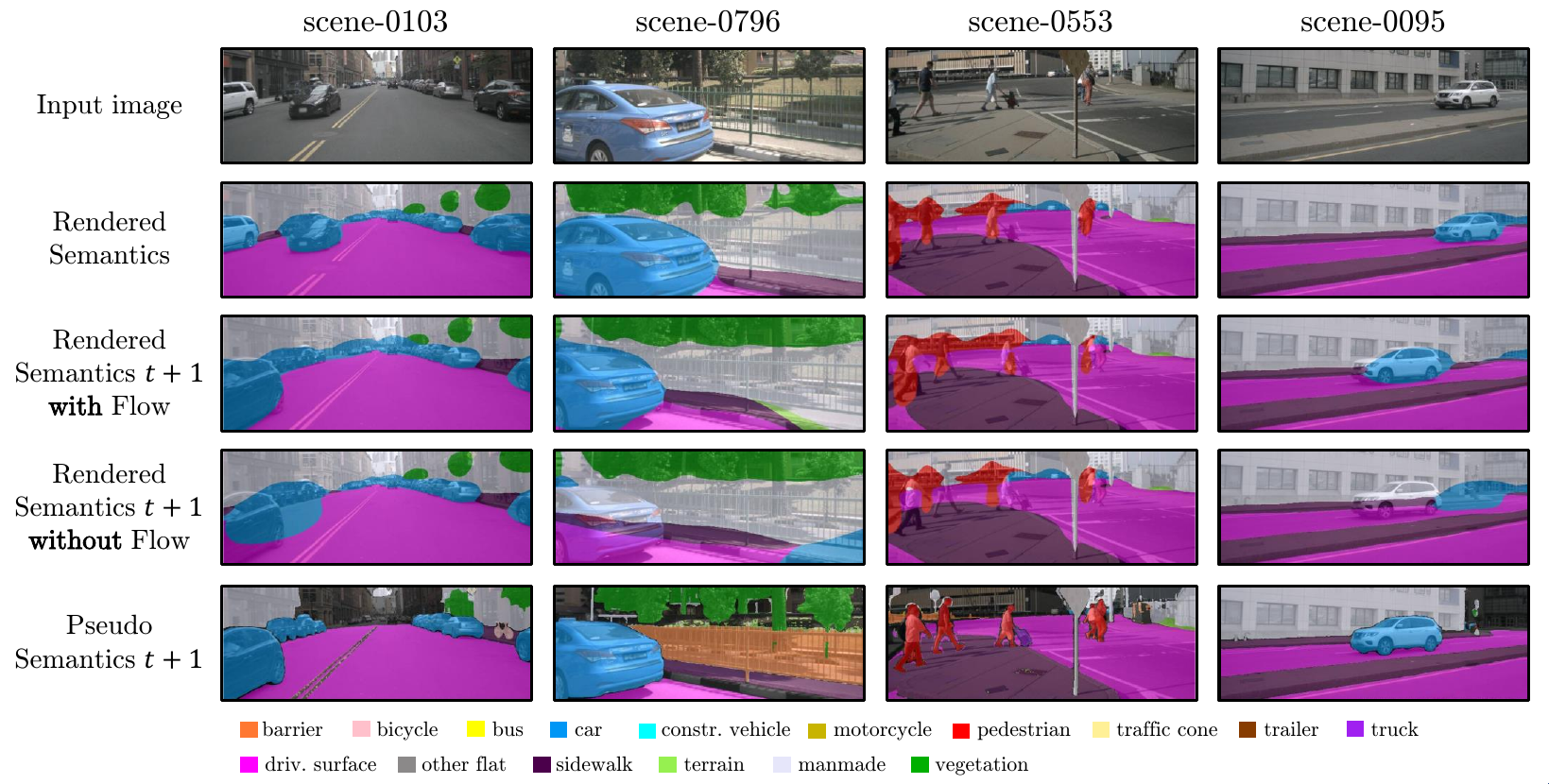}
      \caption{
      \textbf{Qualitative results on the Occ3D-nuScenes validation set.}
      Each column shows an input image and the rendered semantics generated by using GS to render predicted Gaussians into the input camera, as well as rendered semantics when rendering the predictions into the next frame, one with using our \emph{Temporal Module} to compensate motion and one without.
      The last row displays the pseudo labels of the next frame to show how correct motion compensation should look like.}
    \label{fig:qual_render_temporal}
    \end{figure*}
    \clearpage
}

\subsection{Long-term Temporal Flow}
\begin{figure*}
  \centering
      \includegraphics[page=8, trim=0cm 5.4cm 0cm 0cm, clip, width=.99\textwidth]{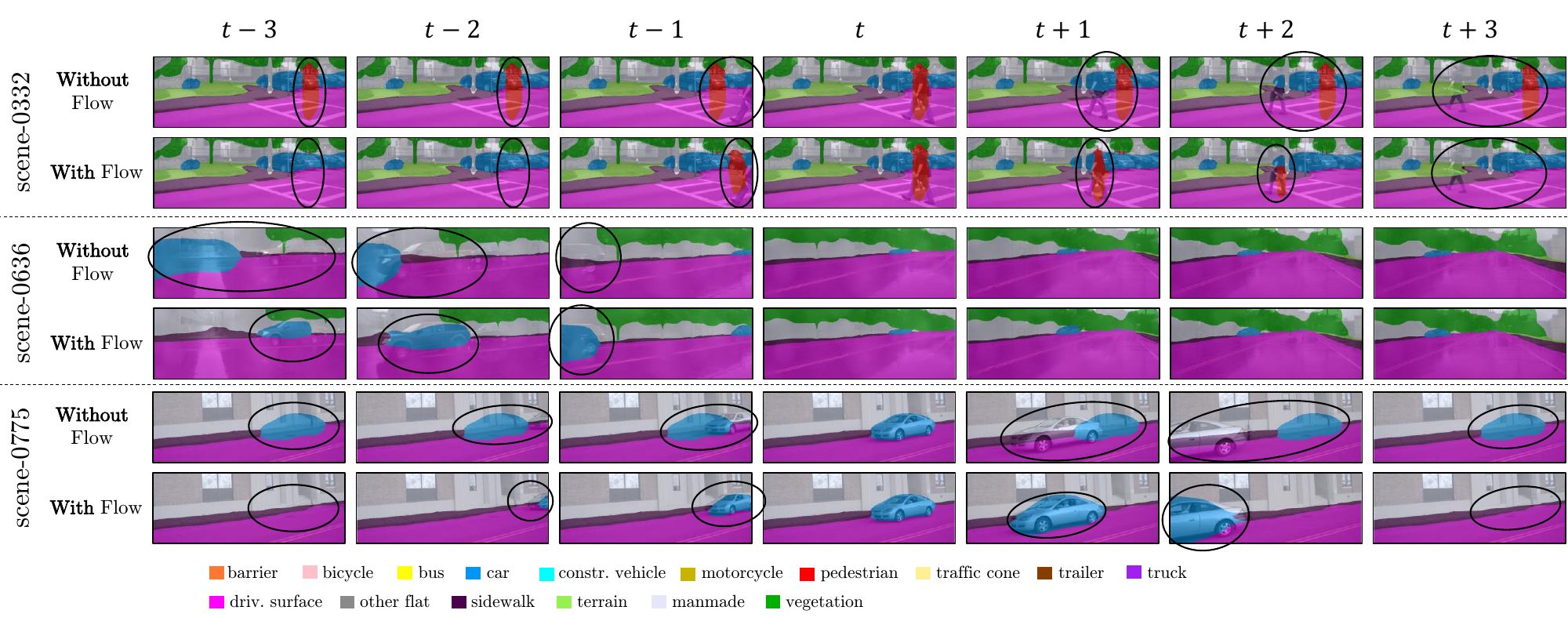}
  \caption{
  \textbf{Qualitative results on the Occ3D-nuScenes validation set.}
    Each row depicts the rendered estimated Gaussians when using the estimated temporal flow to move the Gaussians versus not using the flow, for a time horizon of 1.5s (3 frames) into the past and the future, respectively.
    The Gaussians are predicted for the current frame, and then rendered into adjacent frames.
    It is visible that the model learns to compensate object motion over multiple frames.
    }
\label{fig:qual_render_temporal_horizon}
\end{figure*}

We further examine the behavior of the \emph{Temporal Module} by visualizing the estimated flow over a longer temporal horizon, spanning three frames into the past and future.
\Cref{fig:qual_render_temporal_horizon} displays rendered semantic maps that illustrate the qualitative effectiveness of the estimated 3D flow.
As is always the case, the model predicts the 3D Gaussians for the current frame $t$ given the images of the current frame (along with the previous frame's Gaussians for temporal fusion).
The \emph{Temporal Module} then estimates the motion of objects across each of the surrounding frames, and the Gaussian means are updated according to the predicted flow.
The figure compares rendered results using Gaussians moved by the estimated flow with those using static, unmoved Gaussians.
The visualizations demonstrate that the model learns reasonable object motion that better aligns with the temporal position of the camera frame, despite not being explicitly trained with motion or scene flow supervision.
The \emph{Temporal Module} is optimized solely through consistency constraints between rendered temporal frames and the Gaussians adjusted by the estimated flow.
This temporal consistency reduces artifacts during \emph{Temporal Gaussian Splatting}, which is critical for stable training.
We note that learning scene flow in this weakly supervised manner remains challenging, and errors are still present in practice.
Nevertheless, as shown in \cref{sec:ablation_temp_module}, leveraging the estimated flow to improve temporal consistency leads to a notable increase in occupancy estimation performance.

\subsection{Comparison to GaussianOcc}
\begin{figure*}
  \centering
      \includegraphics[page=9, trim=0cm 4.82cm 10.69cm 0cm, clip, width=.99\textwidth]{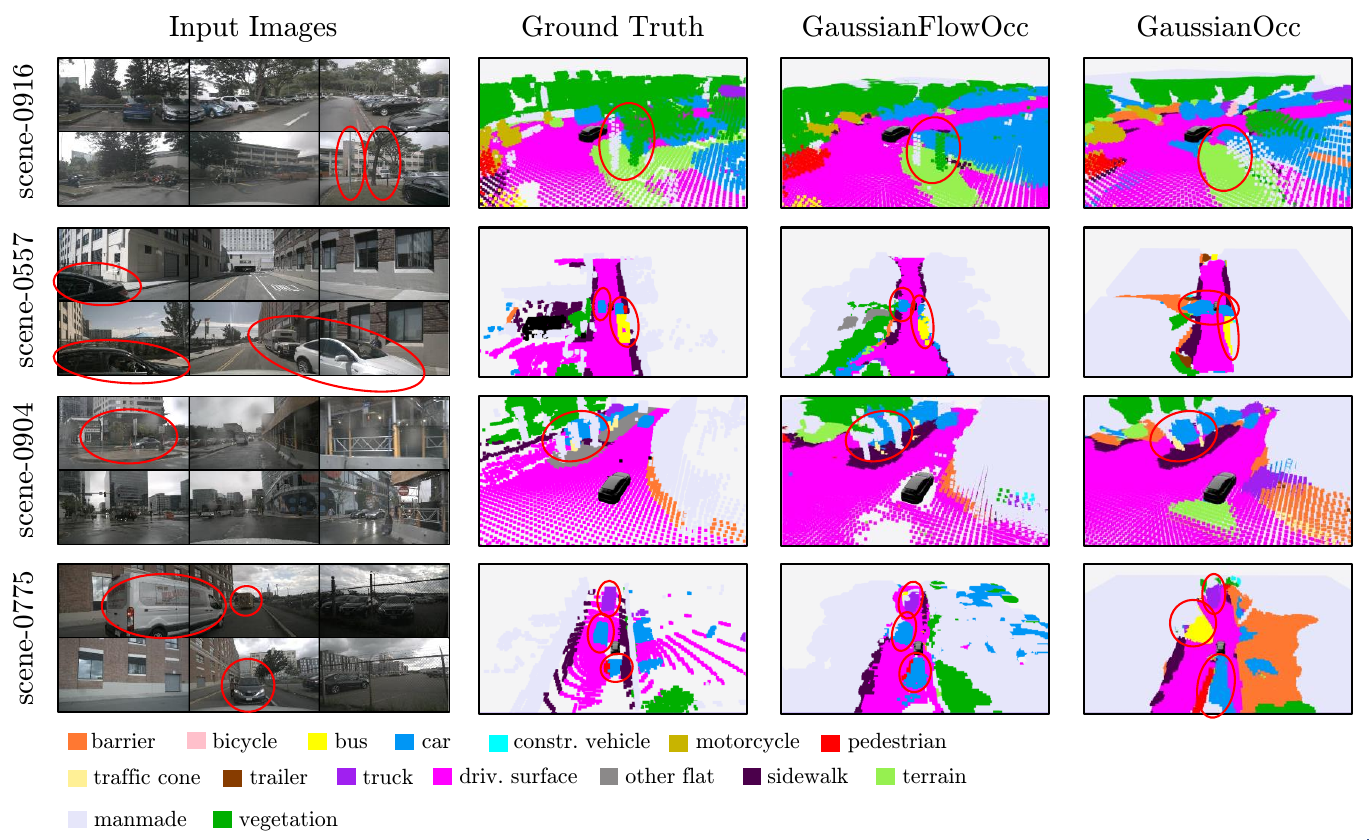}
  \caption{
  \textbf{Comparing our method to GaussianOcc.}
    We compare the occupancy estimation results on four samples of different validation scenes of the nuScenes dataset between our method and GaussianOcc~\cite{gan2024gaussianocc}.
    It is visible that our method can better represent thin and small objects, suffers less from object bleeding and can better estimate the complete 3D shape of scene objects.
    }
\label{fig:qual_comparison}
\end{figure*}

To further demonstrate the benefits of our approach, we compare the predicted occupancy between GaussianFlowOcc and GaussianOcc~\cite{gan2024gaussianocc} in \cref{fig:qual_comparison}.
In contrast to GaussianOcc, our model consistently detects and models thin and small objects like trees, traffic signs and poles (\emph{scene-0916} \& \emph{scene-0904}).
In addition, GaussianFlowOcc suffers a lot less from object bleeding (\emph{scene-0557} \& \emph{scene-0775}).
It is visible that our method can much better estimate the 3D shape of objects (such as vehicles) avoiding the characteristic depth stretching that often affects 2D-supervised methods~\cite{pan2023renderocc, gan2024gaussianocc, boeder2024langocc, huang2023selfocc, zhang2023occnerf}.
We attribute this improvement to our use of long-term temporal supervision, which enforces geometric consistency over time.
This strategy is only viable with an explicit dynamic scene model—highlighting the importance of our integrated motion estimation framework.

\section{Voxelization}
As explained in the main paper, for benchmarking and comparison, we convert the estimated 3D Gaussian distributions into a voxelized representation.
This process begins with defining a voxel grid over the scene.
Each Gaussian distribution is then evaluated at the center of every voxel, where its opacity and semantic logits are accumulated to determine the final voxel values.
The formulation for this voxelization is as follows:
\begin{equation}
    \begin{aligned}
        v_o(p; \mathcal{G}) & = \sum_{i=1}^PG_i(p; \mu_i, s_i, r_i,o_i) \\
        & =  \sum_{i=1}^P \text{exp}\left(- \frac{1}{2}\left( p -\mu_i \right)^T \Sigma^{-1}_i \left( p-\mu_i \right) \right) o_i \\
        \end{aligned}
        \end{equation}
        \begin{equation}
        \begin{aligned}
        v_c(p; \mathcal{G}) & =  \sum_{i=1}^PG_i(p; \mu_i, s_i, r_i,c_i) \\ 
        & = \sum_{i=1}^P \text{exp}\left(- \frac{1}{2}\left( p -\mu_i \right)^T \Sigma^{-1}_i \left( p-\mu_i \right) \right) c_i  ,
    \end{aligned}
\end{equation}
where $\Sigma_i$ represents the covariance matrix of each Gaussian, derived from its rotation quaternion and scale parameters.
To ensure efficiency in handling large numbers of Gaussians and voxels, we limit the influence of each Gaussian to a fixed local neighborhood.
This is justified by the fact that the contribution of each Gaussian typically decays rapidly with distance, making it computationally unnecessary to evaluate its impact on distant voxels.
As explained in \cref{sec:3d_labels}, this voxelization operation is fully differentiable, which enables training of our model using 3D voxel labels when available.

\section{Source Code}\label{sec:source_code}
Source code to reproduce the results presented in the paper is available at \url{https://github.com/boschresearch/GaussianFlowOcc}.
Please follow the instructions in the \textit{README.md} file to install the repository and run the code.
We provide instructions to train and evaluate our models on the Occ3D-nuScenes dataset.

\end{document}